\documentclass[twoside,11pt]{article}

\usepackage{mlhc}
\usepackage{custom_style}
\usepackage{bm}
\usepackage{amsmath}
\usepackage{hyperref}
\usepackage{longtable,lscape}
\usepackage[normalem]{ulem}
\usepackage{booktabs}
\RequirePackage{color}
\definecolor{mydarkblue}{rgb}{0,0.08,0.45}
\hypersetup{ % play with the different link colors here
    colorlinks,
    citecolor=mydarkblue,
    filecolor=mydarkblue,
    linkcolor=mydarkblue,
    urlcolor=mydarkblue % set to black to prevent printing blue links
}
\usepackage{subcaption}
\usepackage{url}
\usepackage{floatrow}
\usepackage[top=1in, bottom=1.25in, left=1in, right=1in]{geometry}

\newcommand{\blue}[1]{\textcolor{blue}{\textbf{#1}}}

\begin{document}

\title{Doctor AI: Predicting Clinical Events\\ via Recurrent Neural Networks}

\author{\name Edward Choi, Mohammad Taha Bahadori \email {mp2893,bahadori}@gatech.edu \\
\addr College of Computing\\
Georgia Institute of Technology\\
Atlanta, GA, USA 
\AND
\name Andy Schuetz, Walter F. Stewart \email {schueta1,stewarwf}@sutterhealth.org \\
\addr Research Development \& Dissemination\\
Sutter Health\\
Walnut Creek, CA, USA
\AND
\name Jimeng Sun \email jsun@cc.gatech.edu \\
\addr College of Computing\\
Georgia Institute of Technology\\
Atlanta, GA, USA
} 

\maketitle

\vspace{-10mm}
\begin{abstract}
 Leveraging large historical data in electronic health record (EHR), we developed Doctor AI, a generic predictive model that covers observed medical conditions and medication uses. Doctor AI is a temporal model using recurrent neural networks (RNN) and was developed and applied to longitudinal time stamped EHR data from 260K patients over 8 years. Encounter records (e.g. diagnosis codes, medication codes or procedure codes) were input to RNN  to predict (all) the diagnosis and medication categories for a subsequent visit. Doctor AI assesses the history of patients to make multilabel predictions (one label for each diagnosis or medication category). Based on separate blind test set evaluation, Doctor AI can perform differential diagnosis with up to 79\% recall@30, significantly higher than several baselines. Moreover, we demonstrate great generalizability of Doctor AI by adapting the resulting models from one institution to another without losing substantial accuracy.
  
\end{abstract}

\vspace{-2mm}
\section{Introduction}
\vspace{-1mm}
A common challenge in healthcare today is that physicians have access to massive amounts of data on patients, but little time nor tools.
Intelligent clinical decision support anticipates the information at the point of care that is specific to the patient and provider needs.  
Electronic health records (EHR), now commonplace in U.S. healthcare, represent the longitudinal experience of both patients and doctors. These data are being used with increasing frequency to predict future events. 
While predictive models have been developed to anticipate needs, most existing work has focused on specialized predictive models that predict a limited set of outcomes. However, day-to-day clinical practice involves an unscheduled and heterogeneous mix of scenarios and needs different prediction models in the hundreds to thousands. It is impractical to develop and deploy specialized models one by one.

Leveraging large historical data in EHR, we developed Doctor AI, a generic predictive model that covers observed medical conditions and medication uses. Doctor AI is a temporal model using recurrent neural networks (RNN) and was developed and applied to longitudinal time stamped EHR data.
In this work, we are particularly interested in whether historical EHR data may be used to predict future physician diagnoses and medication orders. Applications that accurately forecast could have many uses such as anticipating the patient status at the time of visit and presenting data a physician would want to see at the moment. The primary goal of this study was to use longitudinal patient visit records to predict the physician diagnosis and medication order of the next visit. As a secondary goal we predicted the time to the patient’s next visit. Predicting the visit time facilitates guidance of whether a patient may be delayed in seeking care.

The two tasks addressed in this work are different from sequence labeling tasks often seen in natural language processing applications, e.g., part-of-speech tagging. Our proposed model, Doctor AI, performs multilabel prediction (one for each disease or medication category) over time while sequence labeling task predicts a single label at each step. The key challenge was finding a flexible model that is capable of performing the multilabel prediction problem. The two main classes of techniques have been proposed in dealing with temporal sequences: 
1) continuous-time Markov chain based models \citep{nodelman2002continuous, lange2015joint, johnson2013bayesian}, and
2) intensity based point process modeling techniques such as Hawkes processes \citep{liniger2009multivariate, zhu2013nonlinear,choiconstructing2015}. 
However, both classes are expensive to compute, especially for nonlinear settings. Furthermore, they often make strong assumptions about the data generation process which might not be valid for EHR data.
Our modeling strategy was to develop a generalized approach to representing patient temporal healthcare experience to predict all the diagnoses, medication categories and visit time. We used recurrent neural network (RNN), considering that RNNs have been particularly successful for representation learning in sequential data, \textit{e.g.} \cite{graves2013generating,graves2014towards,sutskever2014sequence,kiros2014unifying,zaremba2014learning}. In particular, we make the following main contributions in this paper:
\begin{itemize}
\vspace{-1mm}
\item
We demonstrate how RNNs can be used to represent the patient status and predict diagnosis, medication order and visit time. The trained RNN is able to achieve above 64\% recall@10 and 79\% recall@30 for diagnosis prediction, showing potential to serve as a differential diagnosis assistance.
\vspace{-1mm}
\item
We propose an initialization scheme for RNNs using Skip-gram embeddings \citep{mikolov2013distributed} and show that it improves the performance of the RNN in both accuracy and speed.
\vspace{-1mm}
\item
We empirically confirm that RNN models possess great potential for transfer learning across different medical institutions. This suggests that health systems with insufficient patient data can adopt models learned from larger datasets of other health systems to improve prediction accuracy on their smaller population.
\end{itemize}

\vspace{-4mm}
\section{Related Work}
\vspace{-1mm}
In this section, we briefly review the common approaches to modeling multilabel event sequences with special focus on the models that have been applied to medical data. There are two main approaches to modeling multilabel event sequences: with or without discretization (binning) of time. 

\noindent \textbf{Discretization.}
When the time axis is discretized, the point process data can be converted to binary time series (or time series of count data if binning is coarse) and analyzed via time series analysis techniques \citep{truccolo2005point,bahadori2013fast,ranganathsurvival}. However, this approach is inefficient as it produces long time series whose elements are mostly zero. Furthermore, discretization of time introduces noise in the time stamps of visits. Finally, these approaches are often not able to model the duration until next event. Thus, it is advantageous not to discretize the data both in terms of modeling and computation.

\noindent \textbf{Continuous-time models.}
Among the continuous-time models, there are two main techniques: continuous-time Markov chain based models \citep{foucher2007semi,johnson2013bayesian,lange2014latent,liu2013longitudinal} and their extension using Baysian networks \citep{nodelman2002continuous,weiss2012multiplicative} and intensity function modeling techniques such as Cox and Hawkes processes \citep{liniger2009multivariate,zhou2013learning,linderman2014discovering,choiconstructing2015}. %\ecedit{Continuous-time Bayesian Network models such as \citep{} also deal with temporal data, but they focus on learning the temporal dependencies among variables rather than maximizing the prediction accuracy.}

Intensity function modeling techniques have been shown to have computational advantages over the continuous-time Markov chain based models. Moreover, modeling multilabel marked point processes with continuous-time Markov chains expands their state-space and make them even more expensive.
However, Hawkes processes only depend linearly on the past observation times; while there are limited classes of non-linear Hawkes process \citep{zhu2013nonlinear}, the temporal dynamics can be more complex. %Moreover, there is no scalable multi-label extension for Hawkes processes. 
Finally, Hawkes processes are known to have a flat loss function near optimal value of the parameters which renders the gradient-based learning algorithms inefficient \citep{veen2008estimation}. In this paper we address these challenges by designing a recurrent neural network which has been shown to be successful in learning complex sequential patterns.

\noindent \textbf{Disease progression models.} 
There have been active research in modeling the temporal progression of diseases \citep{mould2012models}. Generally, most works can be divided into two groups: works that focus on a specific disease and works that focus on a broader range of diseases.

\textit{Specific-purpose progression modeling:}
There have been many studies that focus on modeling the temporal progression of a specific disease based on either intensive use of domain-specific knowledge \citep{de2006mechanism,ito2010disease,tangri2011predictive} or taking advantage of advanced statistical methods \citep{liu2013longitudinal,jackson2003multistate,sukkar2012disease,zhou2012modeling}. Specifically, studies have been conducted on Alzheimer's disease \citep{ito2010disease,zhou2012modeling,sukkar2012disease}, glaucoma \citep{liu2013longitudinal}, chronic kidney disease \citep{tangri2011predictive}, diabetes mellitus \citep{de2006mechanism}, and abdominal aortic aneurysm \citep{jackson2003multistate} 

\textit{General-purpose progression modeling:}
Recently, \citet{wang2014unsupervised,choiconstructing2015,ranganathsurvival} proposed more general approaches to modeling the progression of wider range of diseases. As discussed earlier, \citet{choiconstructing2015} used Hawkes process, and \citet{ranganathsurvival} discretized time in order to model multiple patients and multiple diseases. \citet{wang2014unsupervised} proposed a graphical model based on Markov Jump Process to predict the stage progression of chronic obstructive pulmonary disease (COPD) and its co-morbid diseases. 

One of the main challenges in using these algorithms is scalability. The datasets used in previous works typically contain up to a few thousands of patients and a few hundreds of codes. Even the largest dataset used by \citet{ranganathsurvival} contains 13,180 patients and 8,722 codes, which is significantly smaller than our dataset described in Table \ref{tab:basic}. Need for domain-specific knowledge is also a big challenge. For example, \citet{wang2014unsupervised} not only used a smaller dataset (3,705 patients and 264 codes) but also used co-morbidity information to improve the performance of their algorithm. Such expert knowledge is difficult to obtain from typical EHR data. 

\noindent \textbf{Deep learning models for EHR.} Researchers have recently begun attempting to apply neural network based methods (or deep learning) to EHR to utilize its ability to learn complex patterns from data. Previous studies such as phenotype learning \citep{lasko2013computational,che2015deep,hammerla2015pd} or representation learning \citep{choi2016learning,choi2016multi,miotto2016deep}, however, have not fully addressed the sequential nature of EHR. \citet{lipton2016learning} is especially related to our work in that both studies use RNN for sequence prediction. However, while \citet{lipton2016learning} uses regular times series of real-valued variables collected from ICU patients to predict diagnosis codes, we use discrete medical codes (\textit{e.g.} diagnosis, medication, procedure) extracted from longitudinal patient visit records. Also, in each visit we make a prediction about predict diagnosis, medication order in the next visit and and the time to next visit.
\vspace{-0.2in}

\section{Cohort}
\begin{table}[t]
\caption{Basic statistics of the the clinical records dataset.}
\label{tab:basic}
\centering
\begin{tabular}{l|c||l|c}
\# of patients & 263,706 & Total \# of codes & 38,594 \\
\hline
Avg. \# of visits & 54.61 & Total \# of 3-digit Dx codes & 1,183\\
\hline
Avg. \# of codes per visit &3.22 & \# of top level Rx codes & 595\\
\hline
Max \# of codes per visit & 62& Avg. duration between visits & 76.12 days \\
\end{tabular}
\end{table}

\label{sec:cohort}
\noindent \textbf{Population and source of data.} 
The source population for this study was primary care patients from Sutter Health Palo Alto Medical Foundation. Sutter Health is a large primary care and multispecialty group practice that has used an Epic Systems Corporation EHR for more than a decade. The dataset was extracted from a density sampled case-control study for heart failure. The dataset consists of de-identified encounter orders, medication orders, problem list records and procedure orders.

\noindent \textbf{Data processing.} As inputs, we use ICD-9 codes, medication codes, and procedure codes. We extracted ICD-9 codes from encounter records, medication orders, problem list records and procedure orders. Generic Product Identifier (GPI) medication codes and CPT procedure codes were extracted from medication orders and procedure orders respectively. All codes were timestamped with the patients’ visit time. If a patient received multiple codes in a single visit, those codes were given the same timestamp. We excluded patients that made less than two visits. The resulting dataset consists of 263,706 patients who made on average 54.61 visits per person. 

\noindent \textbf{Grouping medical codes.}
There are more about 11,000 unique ICD-9 codes and 18,000 GPI medication codes in the dataset, many of which are very granular. For example, pulmonary tuberculosis (ICD-9 code 011) is divided into 70 subcategories (ICD-9 code 011.01, 011.02, ..., 011.95, 011.96). Simply knowing that a patient is likely to have pulmonary tuberculosis is enough to increase the doctor's awareness of the severity of the clinical situation. Therefore, to predict diagnosis and medication order, we grouped codes into higher-order categories to reduce the feature set and information overload. For the diagnosis codes, we use the 3-digit ICD-9 codes, yielding 1183 unique codes. For the medication codes, we use the Generic Product Identifier Drug Class, which groups the medication codes into 595 unique groups.  The label $\yb_i$ we use in the following sections represents the 1,778-dimensional vector (i.e., 1183 + 595) for the grouped diagnosis codes and medication codes.

\vspace{-2mm}
\section{Methods}
This section describes the RNN model for multilabel point processes.  We will also describe how we predict diagnosis, medication order and visit time using the RNN model.

\noindent \textbf{Problem setting.}
For each patient, the observations are drawn from a  multilabel point process in the form of $(t_i, \bm{x}_i)$ for $i = 1, \ldots, n$. 
Each pair represents an event, such as an ambulatory care visit, during which multiple medical codes such as ICD-9 diagnosis codes, procedure codes, or medication codes are documented in the patient record.  
The multi-hot label vector $\bm{x}_i \in \{0,1\}^p$ represents the medical codes assigned at time $t_i$, where $p$ denotes the number of unique medical codes. 
At each timestamp, we may extract higher-level codes for prediction purposes and denote it by $\bm{y}_i$, see the details in Section \ref{sec:cohort}. 
The number of events for each patient may differ.

\noindent \textbf{Gated Recurrent Units Preliminaries.}
Specifically, we implemented our RNN with Gated Recurrent Units (GRU). Although Long Short Term Memory (LSTM) \citep{hochreiter1997long,graves2009novel} has drawn much attention from many researchers, GRU has recently shown to have similar performance as LSTM, while employing a simpler architecture \citep{chung2014empirical}. In order to precisely describe the network used in this work, we reiterate the mathematical formulation of GRU as follows:
\begin{align*}
\zb_i & = \sigma(\Wb_z \xb_i + \Ub_z \hb_{i-1} + \bb_z) \\
\rb_i & = \sigma(\Wb_r \xb_i + \Ub_r \hb_{i-1} + \bb_r) \\
\tilde{\hb}_i & = \tanh(\Wb_h \xb_i + \rb_i \circ \Ub_h \hb_{i-1} + \bb_h) \\
\hb_i & = \zb_i \circ \hb_{i-1} + (1 - \zb_i) \circ \tilde{\hb}_i
\end{align*}
where $\zb_i$ and $\rb_i$ respectively represent the update gate and the reset gate, $\tilde{\hb}_i$ the intermediate memory unit, $\hb_i$ the hidden layer, all at timestep $t_i$. A detailed description of GRU is provided in Appendix \ref{sec:gru}.

\noindent \textbf{Description of neural network architecture.}
Our goal is to learn an effective vector representation for the patient status at each timestamp $t_i$. Using effective patient representations, we are interested in predicting diagnosis and medication categories in the next visit $\bm{y}_{i+1}$ and the time duration until the next visit $d_{i+1}=t_{i+1}-t_i$. Finally, we would like to perform all these steps jointly in a single supervised learning scheme. We use RNN to learn such patient representations, treating the hidden layer as the representation for the patient status and use it for the prediction tasks.

%\sout{To learn such patient representations we use recurrent neural networks, because patients have different length of medical records and also recurrent neural networks have been shown to be particularly successful for representation learning in sequential data, e.g.}. \sout{We will take the state vector of RNNs as the latent representation for patients and use it for predicting multiple forms of outputs.}

\begin{figure}[t]
\floatbox[{\capbeside\thisfloatsetup{capbesideposition={left,center},capbesidewidth=10cm}}]{figure}[\FBwidth]
{\caption{This diagram shows how we have applied RNNs to solve the problem of forecasting of next visits' time and the codes assigned during each visit. The first layer simply embeds the high-dimensional input vectors in a lower dimensional space. The next layers are the recurrent units (here two layers), which learn the status of the patient at each timestamp as a real-valued vector. Given the status vector, we use two dense layers to generate the codes observed in the next timestamp and the duration until next visit.}\label{fig:highlevel}}
{\qquad\includegraphics[scale=0.3]{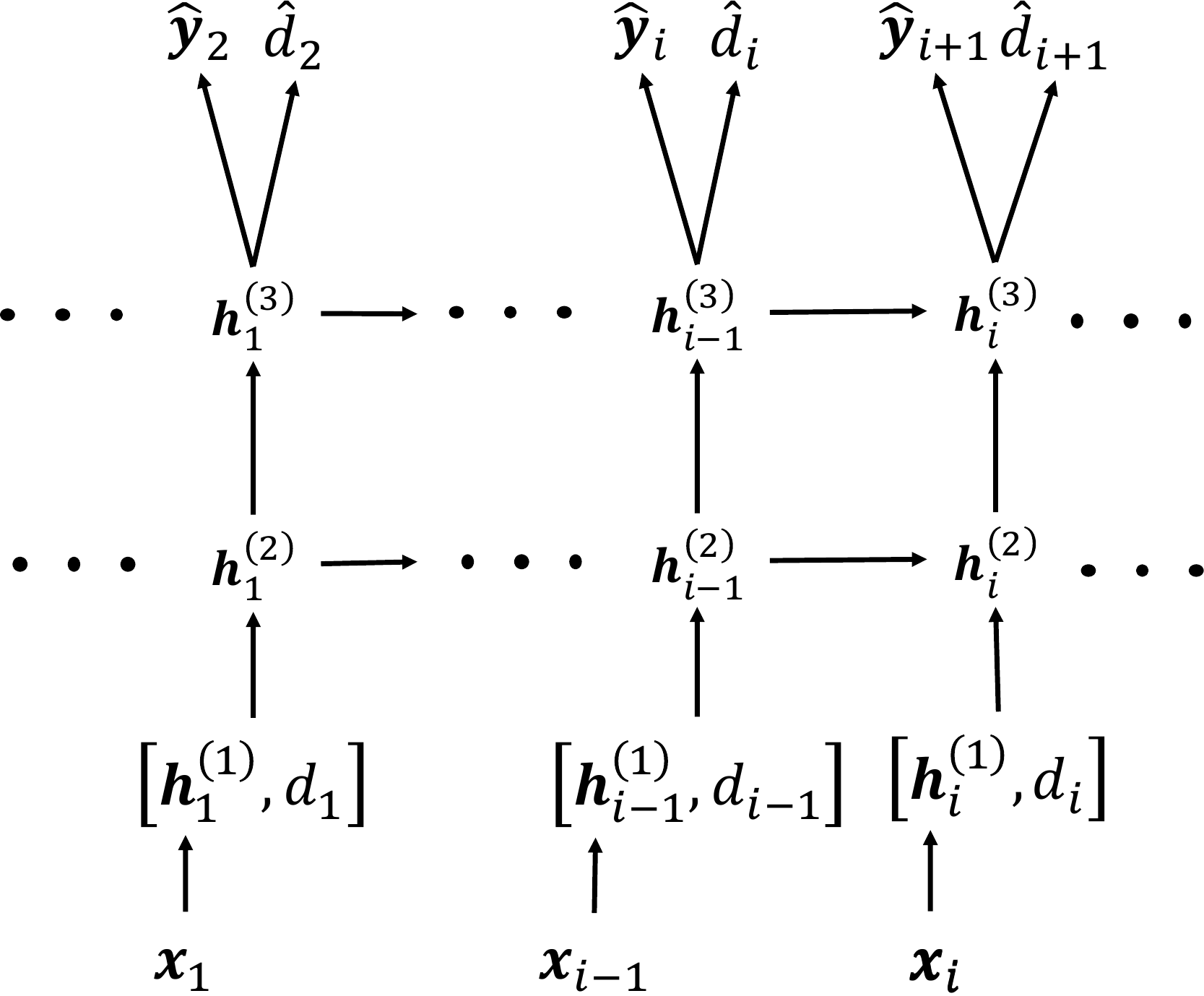}}   
\end{figure}

The proposed neural network architecture (Figure \ref{fig:highlevel}) receives input at each timestamp $t_i$ as the concatenation of the multi-hot input vector $\bm{x}_i$ of the multilabel categories and the duration $d_{i}$ since the last event. In our datasets, the input dimension is as large as $40,000$. Thus, the next layer projects the input to a lower dimensional space. Then, we pass the lower dimensional vector through RNN (implemented with GRU in our study). We can also stack multiple layers of RNN to increase the representative power of the network. Finally, we use a Softmax layer to predict the diagnosis codes and the medication codes, and a rectified linear unit (ReLU) to predict the time duration until next visit.

% \begin{wrapfigure}{r}{0.5\textwidth}
% \centering
% \includegraphics[scale=0.3]{./Figs/rnn_diagram}
% \vspace{0.4in}
% \caption{This diagram shows how we have applied RNNs to solve the problem of joint forecasting of next visits' time and the codes assigned during each visit. The first layer simply embeds the high-dimensional input vectors in a lower dimensional space. The next layers are the recurrent units (here two layers) followed by a dense layer to generate each output type.}
% \label{fig:highlevel}
% \vspace{-0.65in}
% \end{wrapfigure}

% \begin{figure}[t]
% \centering
% \includegraphics[scale=0.3]{./Figs/rnn_diagram}
% \caption{This diagram shows how we have applied RNNs to solve the problem of joint forecasting of next visits' time and the codes assigned during each visit. The first layer simply embeds the high-dimensional input vectors in a lower dimensional space. The next layers are the recurrent units (here two layers) followed by a dense layer to generate each output type.}
% \label{fig:highlevel}
% \end{figure}

For predicting the diagnosis codes and the medication codes at each timestep $t_i$, a Softmax layer is stacked on top of the GRU, using the hidden layer $\hb_i$ as the input: $\widehat{\yb}_{i+1} = \mathrm{softmax}({\Wb_{code}}^{\top} \hb_i + \bb_{code})$.
% , which can be formulated as follows:
% \begin{align*}
% \widehat{\yb}_{t+1} = \frac{\exp({\Wb_{code}}^{\top} \hb_i + \bb_{code})}{\sum{\exp({\Wb_{code}}^{\top} \hb_i + \bb_{code})}}
% \end{align*}
% where $\Wb_{code}$ is the weight used to predict the future codes.
For predicting the time duration until the next visit, a rectified linear unit (ReLU) is placed on top of the GRU, again using the hidden layer $\hb_i$ as the input: $\widehat{d}_{i+1} = \max({\wb_{time}}^{\top} \hb_i + b_{time}, 0)$.
% , which can be formulated as follows:
% \begin{align*}
% \widehat{d}_{t+1} = \max({\wb_{time}}^{\top} \hb_i + b_{time}, 0)
% \end{align*}
The objective of training our model is to learn the weights $\Wb_{\{z,r,h,code\}}$, $\Ub_{\{z,r,h\}}$, $\bb_{\{z,r,h,code\}}$, $\wb_{time}$ and $b_{time}$. The values of all $\Wb$'s and $\Ub$'s were initialized to orthonormal matrices using singular value decomposition of matrices generated from the normal distribution \citep{saxe2013exact}. The initial value of $\wb_{time}$ was chosen from the uniform distribution between $-0.1$ and $0.1$. All $\bb$'s and $b_{time}$ were initialized to zeros. The joint loss function consists of the cross entropy for the code prediction and the squared loss for the time duration prediction, as described below for a single patient:
\begin{align*}
\mathcal{L}(\Wb, \Ub, \bb, \wb_{time}, b_{time}) = \sum_{i=1}^{n-1} \bigg\{ \Big( \yb_{i+1} \log(\widehat{\yb}_{i+1}) + (1 - \yb_{i+1})  \log(1 - \widehat{\yb}_{i+1})  \Big) + \frac{1}{2} \|d_{i+1} - \widehat{d}_{i+1}\|_2^2 \bigg\}
\end{align*} 
As mentioned above, the multi-hot vectors $\xb_i$ of almost 40,000 dimensions are first projected to a lower dimensional space, then put into the GRU. 
We employed two different approaches for this: (1) We put an extra layer of a certain size between the multi-hot input $\xb_i$ and the GRU, and call it the embedding layer.
We denote the weight matrix between the multi-hot input vector and the embedding layer as $\Wb_{emb}$.
Then we learn the weight $\Wb_{emb}$ as we train the entire model. 
(2) We initialize the weight $\Wb_{emb}$ with a matrix generated by Skip-gram algorithm \citep{mikolov2013distributed}, then refine the weight $\Wb_{emb}$ as we train the entire model. This can be seen as using the pre-trained Skip-gram vectors as the input to the RNN and fine-tuning them with the joint prediction task. The brief description of learning the Skip-gram vectors from the EHR is provided in Appendix \ref{sec:skipgram}.
The first and second approach can be formulated as follows:
\begin{align}
\bm{h}^{(1)}_i & = [\tanh({\xb_i}^{\top} \Wb_{emb} + \bb_{emb}),~ d_i] \label{eq:embedding}\\
  \bm{h}^{(1)}_i & = [{\xb_i}^{\top} \Wb_{emb},~ d_i]  \label{eq:skipgram}
\end{align}
where $[\cdot,\cdot]$ is the concatenation operation used for appending the time duration to the multi-hot vector $\bm{h}^{(1)}_i$ to make it an input vector to the GRU.  

\vspace{-2mm}
\section{Results}
We now describe the details of our experiments in the proposed RNN approach to forecasting the future clinical events. The source code of Doctor AI is publicly available at \url{https://github.com/mp2893/doctorai}.

\vspace{-2mm}
\subsection{Experiment Setup} 
For training all models including the baselines, we used 85\% of the patients as the training set and 15\% as the test set. We trained the RNN models for 20 epochs (\textit{i.e.}, 20 iterations over the entire training data) and then evaluated the final performance against the test set. To avoid overfitting, we used dropout between the GRU layer and the prediction layer (\textit{i.e.} code prediction and time duration prediction). Dropout was also used between GRU layers if we were using a multi-layer GRU. We also applied norm-$2$ regularization on both $\Wb_{code}$ and $\wb_{time}$. Both regularization coefficients were set to 0.001. The size of the hidden layer $\hb_i$ of the GRU was set to 2000 to guarantee a sufficient expressive power. After running sets of preliminary experiments where we tried the size from 100 to 2000, we noticed that the code prediction performance started to saturate around 1600$\sim$1800. All models were implemented with Theano \citep{Bastien-Theano-2012} and trained on a machine equipped with two Nvidia Tesla K80 GPUs.

We train total four different variation of Doctor AI as follows,
\begin{itemize}
\item
{\bf RNN-1}: RNN with a single hidden layer  initialized with a random orthogonal matrix for $\Wb_{emb}$.
\item
{\bf RNN-2}:  RNN with two hidden layers initialized with a random orthogonal matrix for $\Wb_{emb}$.
\item
{\bf RNN-1-IR}: RNN using a single hidden layer initialized embedding matrix $\Wb_{emb}$ with the Skip-gram vectors trained on the entire dataset.
\item
{\bf RNN-2-IR}: RNN with two hidden layers initialized embedding matrix $\Wb_{emb}$ with the Skip-gram vectors trained on the entire dataset.
 dataset.
\end{itemize}

\vspace{-3mm}
\subsection{Evaluation metrics}
The performance of algorithms in predicting diagnoses and medication codes was evaluated using the Top-k recall defined as:
\begin{equation*}
\text{top-$k$ recall} = \frac{\text{\# of true positives in the top $k$ predictions} }{ \text{ \# of true positives } }
\end{equation*}
Top-k recall mimics the behavior of doctors conducting differential diagnosis, where doctors list most probable diagnoses and treat patients accordingly to identify the patient status. Therefore, a machine with a high Top-k recall translates to a doctor with an effective diagnostic skill. This makes Top-k recall an attractive performance metric for our problem.

We select the maximum k to be 30 to evaluate the performance of the models not only for simple cases but also for complex cases. Near 50.7\% of the patients have been assigned with more than 10 diagnosis and medication codes at least once. Since it is those complex cases that are of interest to predict and analyze, we choose k to be large enough (i.e., 3 times of the mean).

\textbf{Coefficient of determination ($R^2)$} was used to evaluate the predictive performance of regression and forecasting algorithms. It compares the accuracy of the prediction with respect to the simple prediction by mean of the target variable.
\begin{equation*}
R^2 = 1 - \frac{\sum_{i}\left(y_i - \widehat{y_i}\right)^2}{\sum_{i}\left(y_i - \overline{y_i}\right)^2}
\end{equation*}
Because time to the next visit can be highly skewed, we measure the $R^2$ performance of the algorithms in predicting $\log(d_i)$ to lower the impact of anomalous long durations in the performance metric. In the same spirit, we train all models to predict the logarithm of the time duration between visits.

\vspace{-2mm}
\subsection{Baselines}
We compare our model against several baselines as described below. Some of the existing techniques based on continuous-time Markov chain and latent space models were not scalable enough to be trained using the entire dataset in a reasonable amount of time; thus comparison is not feasible.

\noindent \textbf{Frequency baselines.}
We compare our algorithms against simple baselines that are based on experts' intuition about the dynamics of events in clinical settings. The first baseline uses a patient's medical codes in the last visit as the prediction for the current visit. This baseline is competitive when the status of a patient with a chronic condition stabilizes over time. We enhanced this baseline using the top-$k$ most frequent labels observed in visits prior to the current visits. In the experiments we observe that the baseline of top-$k$ most frequent labels is quite competitive.

\noindent \textbf{Logistic and Neural Network time series models.} 
A common way to perform prediction task is to use $\xb_{i-1}$ to predict the codes in the next visit $\xb_i$ using logistic regression or multilayer perceptron (MLP). To enhance the baseline further, we can use the data from $L$ time lags before and aggregate them $\xb_{i-1} + \xb_{i-2} + …, +\xb_{i-L}$ for some duration $L$ to create the features for prediction of $\xb_i$. Similarly, we can have a model that predicts the time until next visit using rectified linear units (ReLU) as the output activation. We set the lag $L = 5$ so that the logistic regression and MLP can use information from maximum five past visits. The details of MLP design are described in Appendix \ref{sec:base}.

%\noindent \textbf{Manual feature extraction.} 
%We borrow a common feature extracting technique used in information retrieval to improve the performance of learning algorithms. We define a set of \textit{tf-idf} like features \citep{Manning2008} based on the history of patients to not only provide potentially better features, but also give learning algorithms access to longer past information. The \textit{idf} vector is calculated based on the entire dataset. To calculate the \textit{tf} for each patient at each visit, we find the term frequencies in all visits until the current visit. We also generate five values based on the duration information: last, min, max, mean, and std of durations until the current visit. We concatenate the \textit{tf-idf} vector and the duration features with the multihot vector of the current visit and use it in the learning algorithms, i.e., multilayer perceptron in our experiments. Further details of the feature extraction process are provided in Appendix \ref{sec:manFeature}.

\vspace{-2mm}
\subsection{Prediction performance}
Table \ref{tab:mainResults} compares the results of different algorithms with RNN based Doctor AI. We report the results in three settings: when we are interested in (1) predicting only diagnosis codes (Dx), (2) predicting only medication codes (Rx), and (3) jointly predicting Dx codes, Rx codes, and the time duration to next visit. The results confirm that the proposed approach is able to outperform the baseline algorithms by a large margin. Note that the recall values for the joint task are lower than those for Dx code prediction or Rx code prediction because the hypothesis space is larger for the joint prediction task.
% \begin{table}[ht]   
% \centering
% \caption{Accuracy of algorithms in forecasting the future medical activities.}
% \label{tab:mainResults}
% {\renewcommand{\arraystretch}{1.1}
% \begin{tabular}{l|c|c|c|c}
%  & \multicolumn{3}{c|}{Percentage Recall @$k$} &\% $R^2$\\
% Algorithms & $k=10$& $k=20$&$k=30$ & $\log(\Delta t)$\\
% \hline
% \hline
% Previous visit's codes & \multicolumn{3}{|c|}{26.25} & -- \\
% Most frequent past codes & 48.11 & 60.23 & 66.00 & -- \\
% Logistic &17.66 & 26.12 & 31.23& 0.0013 \\
% MLP & 19.49& 30.80& 38.13& 0.0017\\
% Hawkes  & & & & \\
% \hline
% RNN-1 (Dx) & 63.12 & 73.11 & 78.49 & e20 \\
% RNN-2 (Dx) & 63.32 & 73.32 & 78.71 & e20 \\
% %RNN-3 (Dx) & 60.96 & 70.99 & 76.59 & e20,h1000\\
% \hline 
% RNN-1-IR (Dx) & 63.24 & 73.33 & 78.73 & e20 \\
% RNN-2-IR (Dx) & 64.30 & 74.31 & 79.58 & e20 \\
% %RNN-3-IR (Dx) & 64.07 & 74.08 & 79.40 & e20 \\
% \hline
% RNN-1 (Rx) & & & & \\
% RNN-2 (Rx) & & & & \\
% \hline
% RNN-1-IR (Rx) & & & & \\
% RNN-2-IR (Rx) & & & & \\
% \hline
% RNN-1 (Time) & & & & \\
% RNN-2 (Time) & & & & \\
% \hline
% RNN-1-IR (Time) & & & & 0.2539, e20 \\
% RNN-2-IR (Time) & & & & \\
% \hline
% RNN-1 (Dx,Rx,Time) & 53.86 & 65.10 & 71.24 & 0.2519, e20 \\
% RNN-2 (Dx,Rx,Time) & 53.61 & 64.93 & 71.14 & 0.2528 \\
% %RNN-3 (Dx,Rx,Time) & & & & \\
% \hline
% RNN-1-IR (Dx,Rx,Time) & 54.37 & 65.68 & 71.85 & 0.2492, e20 \\
% RNN-2-IR (Dx,Rx,Time) & 54.96 & 66.31 & 72.48 & 0.2534, e20 \\
% %RNN-3-IR (Dx,Rx,Time) & & & & \\
% \hline
% \end{tabular}
% }
% \end{table}

\begin{table}[t]   
\centering
% \begin{center}
\caption{Accuracy of algorithms in forecasting future medical activities. Embedding matrices $\Wb_{emb}$ of both RNN-1 (using one hidden layer) and RNN-2 (using two hidden layers) are initialized with random orthogonal vectors. Embedding matrices $\Wb_{emb}$ of both RNN-1-IR (using one hidden layer) and RNN-2-IR (using two hidden layers) are initialized with Skip-gram vectors trained on the entire dataset.}
\label{tab:mainResults}
{\renewcommand{\arraystretch}{1.15}
\begin{tabular}{@{}l@{\;}||@{\;\;}c@{\;}|@{\;}c@{\;}|@{\;}c@{\;\;}||@{\;}c@{\;}|@{\;}c@{\;}|@{\;}c||@{\;}c@{\;}|@{\;}c@{\;}|@{\;}c@{\;}|@{\;\;}c@{}}
 & \multicolumn{3}{@{\;}c@{\;}||@{\;}}{Dx Only Recall @$k$} & \multicolumn{3}{@{\;}c@{\;}||@{\;}}{Rx Only Recall @$k$} & \multicolumn{4}{c}{Dx,Rx,Time Recall @$k$}\\
Algorithms & $k=10$& $k=20$&$k=30$ & $k=10$& $k=20$&$k=30$ & $k=10$& $k=20$&$k=30$& $R^2$ \\
\hline
\hline
Last visit & \multicolumn{3}{@{\;}c@{\;}||@{\;}}{29.17} & \multicolumn{3}{@{\;}c@{\;}||@{\;}}{13.81} &\multicolumn{3}{c}{26.25}& ---\\
Most freq.  &56.63 &67.39 &71.68 & 62.99 &69.02 &70.07 & 48.11 & 60.23 & 66.00& --- \\
Logistic & 43.24 & 54.04 & 60.76 & 45.80 & 60.02 & 68.93 & 36.04 & 46.32 & 52.53 & 0.0726 \\
MLP & 46.66 & 57.38 & 64.03 & 47.62 & 61.72 & 70.92 & 38.82 & 49.09 & 55.74 & 0.1221 \\
%Feature Ext. &26.12 &39.33 &48.20 &32.27 &51.12 &61.65 &19.60 & 30.80 & 38.13 & 0.0022 \\
%Hawkes  & & & & & & & & & & ---\\
\hline
RNN-1 & 63.12 & 73.11 & 78.49 & 67.99 & 79.55 & \textbf{85.53} & 53.86 & 65.10 & 71.24 & 0.2519\\
RNN-2 & 63.32 & 73.32 & 78.71 & 67.87 & 79.47 & 85.43 & 53.61 & 64.93 & 71.14 & 0.2528\\
\hline
RNN-1-IR & 63.24 & 73.33 & 78.73 & \textbf{68.31} & \textbf{79.77} & 85.52 & 54.37 & 65.68 & 71.85 & 0.2492 \\
%RNN-1-IR (d1000) & & & & & & & & & & \\
RNN-2-IR & \textbf{64.30} & \textbf{74.31} & \textbf{79.58} & 68.16 & 79.74 & 85.48 & \textbf{54.96} & \textbf{66.31} & \textbf{72.48} & \textbf{0.2534}\\
\hline
\end{tabular}
}
\end{table}
% 18.46 &27.55 & 32.71  for logistic all
The superior performance of RNN based approaches can be attributed to the efficient representation that they learn for patients at each visit \citep{bengio2013representation,schmidhuber2015deep}. RNNs are able to learn succinct feature representations of patients by accumulating the relevant information from their history and the current set of codes, which outperformed hand-picked features of frequency baselines.

Table \ref{tab:mainResults} confirms that learning patient representation with RNN is easier with the input vectors that are already efficient representations of the medical codes. The RNN trained with the Skip-gram vectors (denoted by RNN-IR) consistently outperforms the RNN that learns the weight matrix $\Wb_{emb}$ directly from the data, with only one exception, the medication prediction Recall@30, although the differences are insignificant. The results also confirm that having multiple layers when using RNN improves its ability to learn more efficient representations. The results also indicate that a single layer RNN might have enough representative power to capture the dynamics of medications, and adding more layers may not improve the performance.

The results also indicate that our approach significantly improves the accuracy of predicting the time duration until the next visit compared to the baselines. However, the absolute value of $R^2$ metric shows that accurate prediction of time intervals remains as a challenge. We believe achieving significantly better time prediction without extra features should be difficult because the timing of a clinical visit can be affected by many personal factors such as financial status, location of residence, means of transportation, and life style, to name a few. Thus, without such sensitive personal information, which is rarely included in the EHR, accurate prediction of time intervals should be unlikely.

\vspace{-2mm}
\subsection{Understanding the behavior of the network}
\begin{figure}
    \centering
    \begin{subfigure}[b]{0.4\textwidth}
        \includegraphics[width=\textwidth]{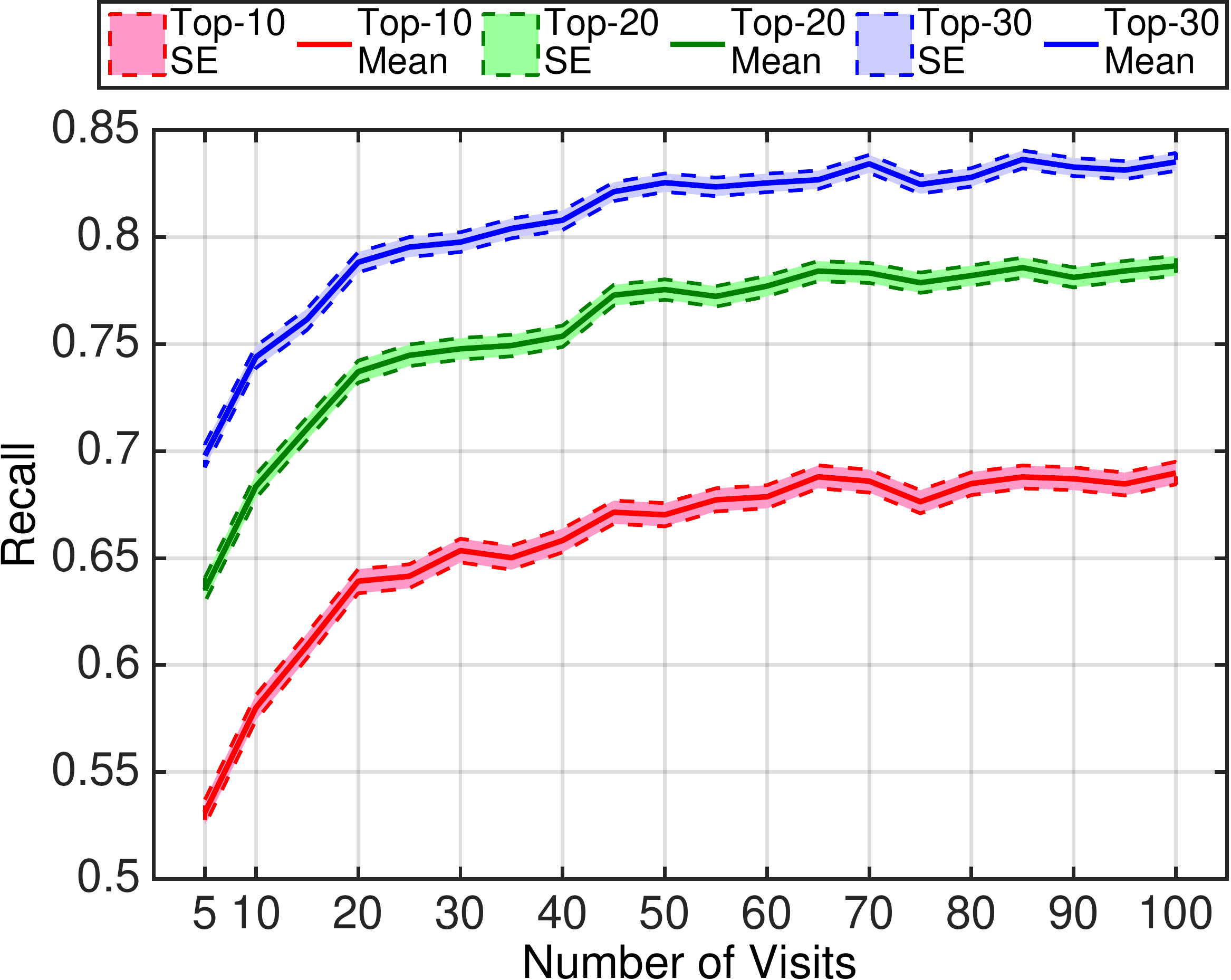}
        \caption{ }
        \label{fig:overTime}
    \end{subfigure}
    \quad %add desired spacing between images, e. g. ~, \quad, \qquad, \hfill etc. 
    \begin{subfigure}[b]{0.4\textwidth}
        \includegraphics[width=\textwidth]{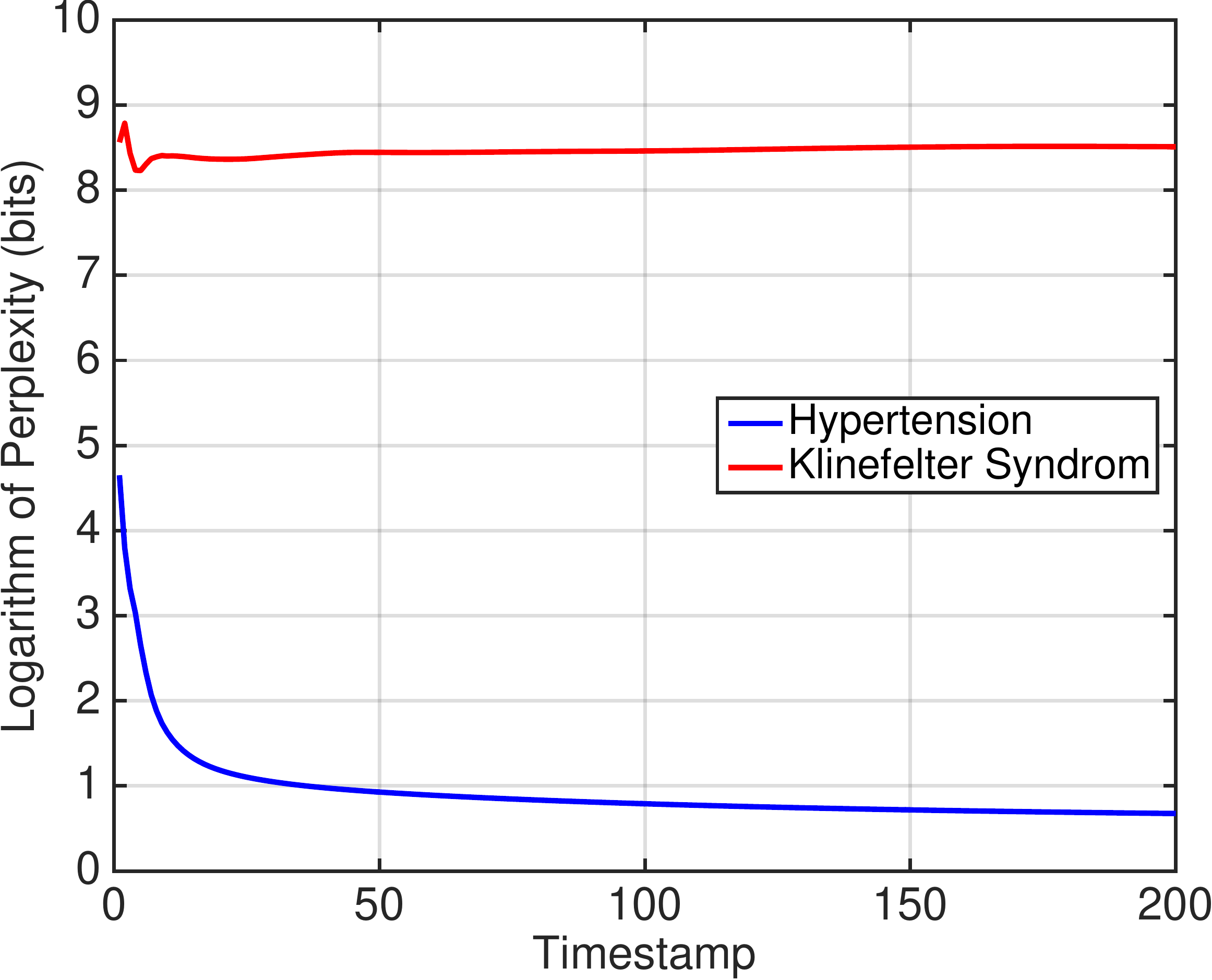}
        \caption{ }
        \label{fig:perplexity}
    \end{subfigure}
    \vspace{-1mm}
    \caption{Characterizing behavior of the trained network: (\subref{fig:overTime}) Prediction performance of Doctor AI as it sees a longer history of the patients.   (\subref{fig:perplexity}) Change in the perplexity of response to a frequent code (hypertension) and an infrequent code (Klinefelter's syndrome).}
    \label{fig:visualization}
\end{figure}

To study the applicability of our model in a real-world setting where patients have varying length of medical records, we conducted an additional experiment to study the relationship between the length of the patient medical history and the prediction performance. To this end, we selected 5,800 patients from the test set who had more than 100 visits. We used the best performing model to predict the diagnosis codes at visits at different times and found the mean and standard error of recall across the selected patients. Figure \ref{fig:overTime} shows the result of the experiment. We believe that the increase in performance can be due to two reasons: (1) RNN is able to learn a better estimate of the patient status as it sees longer patient records and (2) Visits are correlated with poor health. Those with high visit count are more likely to be severely ill, and therefore their future is easier to predict.
%the patient's status stabilizes over time and the prediction task becomes easier.

Another experiment was conducted to understand the behavior of the network by giving synthetic inputs. We chose hypertension (ICD-9 code 401.9) as an example of a frequently observed diagnosis, and Klinefelter's syndrome (ICD-9 code 758.7) as an example of an infrequent diagnosis. We created two synthetic patients who respectively have 200 visits of 401.9 and 758.7. Then we used the best performing model to predict the diagnosis codes for the next visits. 
Figure \ref{fig:perplexity} shows contrasting patterns: when the input is one of the frequent codes such as hypertension, the network quickly learns a more specific set of output codes as next disease. When we select an infrequent code like Klinefelter's syndrome as the input, the network's output is more diverse and mostly the frequently observed codes. The top 30 codes after convergence shown in Table \ref{tab:visu} in Appendix \ref{sec:qual} confirm the disparity of the diversity of the predicted codes for the two cases.

\begin{figure}[t]
\floatbox[{\capbeside\thisfloatsetup{capbesideposition={left,center},capbesidewidth=7cm}}]{figure}[\FBwidth]	
{\caption{The impact of pre-training on improving the performance on smaller datasets. In the first experiment, we first train the model on a small dataset (red curve). In the second experiment, we pre-train the model on our large dataset and use it for initializing the training of the smaller dataset. This procedure results in more than 10\% improvement in the performance. }\label{fig:transfer}}
{\quad\includegraphics[scale=0.28]{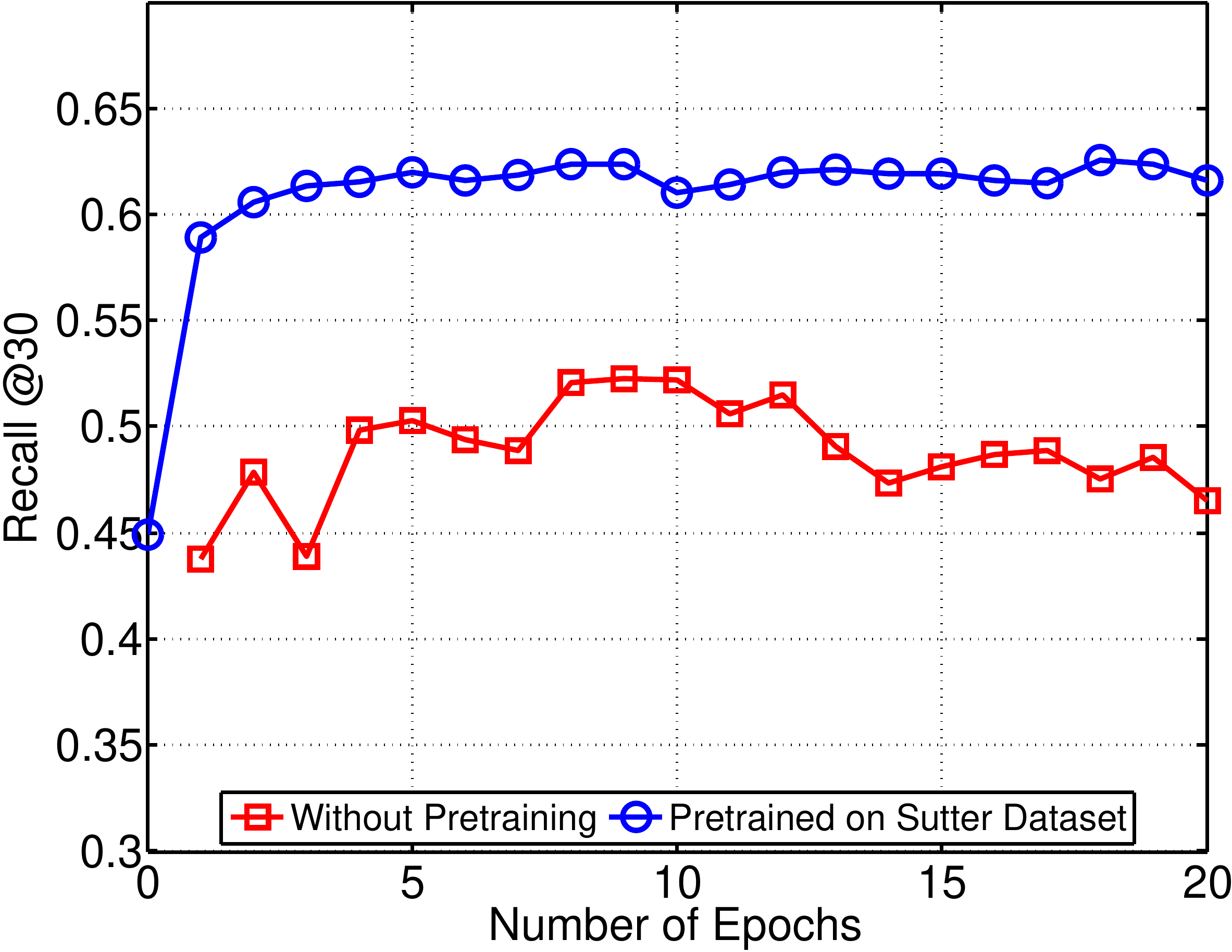}}  
\end{figure}

% \begin{figure}[t]
% \floatbox[{\capbeside\thisfloatsetup{capbesideposition={left,center},capbesidewidth=7cm}}]{figure}[\FBwidth]
% {\caption{Characterizing the perplexity of response for different types of input codes.  When the input is one of the frequent codes such as hypertension, the network quickly learns a more specific set of output codes as next disease.  When we select an infrequent code like Klinefelter Syndrom as the input, the network's output is more diverse and mostly the frequently observed codes. }\label{fig:preplexity}}
% {\quad\includegraphics[scale=0.3]{./Figs/preplexity}}   
% \vspace{-0.3in}
% \end{figure}

\vspace{-2mm}
\subsection{Knowledge transfer across hospitals}
As we observed from the previous experiments, the dynamics of clinical events are complex, which requires models with a high representative power. However, many institutions have not yet collected large scale datasets, and training such models could easily lead to overfitting. To address this challenge, we resort to the recent advances in domain adaptation techniques for deep neural networks  \citep{mesnil2012unsupervised,bengio2012deep,yosinski2014transferable,hoffman2014lsda}.

A different dataset, MIMIC II, which is a publicly available clinical dataset collected from ICU patients over 7 years of observation, was chosen to conduct the experiment. This dataset differs from the Sutter dataset in that it consists of demographically and diagnostically different patients. The number of patients who made at least two visits is 2,695, and the number of unique diagnosis code (3-digit ICD-9 code) is 767, which is a subset of the Sutter dataset. From the dataset, we extracted sequences of 3-digit ICD-9 codes. We chose 2,290 patients for training, 405 for testing. We chose the 2-layer RNN with 1000 dimensional hidden layer, and performed two experiments: 1) We trained the model only on the MIMIC II dataset. 2) We initialized the coefficients of the model with the values learned from the 3-digit ICD-9 sequences of the Sutter data, then we refined the coefficients with the MIMIC II dataset. Figure \ref{fig:transfer} shows the vast improvement of the prediction performance induced by the knowledge transfer from the Sutter data.

%A clinical dataset of 7,653 patients over 5 years of observation from another hospital was chosen to conduct the experiment. This dataset differs from the Sutter dataset in that it consists of demographically and diagnostically different patients. The number of unique diagnosis code in this dataset is 1092, which is a subset of Sutter dataset. From the dataset, we extracted sequences of 3-digit ICD-9 codes. We chose 5,000 patients for training, 2,683 for testing. We chose the 2-layer RNN with 1000 dimensional hidden layer, and performed two experiments. First, we trained the model only on the target dataset. Second, we initialized the coefficients of the model with the values learned from the 3-digit ICD-9 sequences of Sutter data, then we refined the coefficients with the target dataset. Figure \ref{fig:transfer} shows the vast improvement of the prediction performance induced by the knowledge transfer from the Sutter data. It is interesting that the model trained on the Sutter dataset without any refinement achieves a significantly higher recall, and further refinements improve the recall up to 15\%.

% \section{Discussion}
% \input{discuss.tex}

\vspace{-2mm}
\section{Conclusion}
In this work, we proposed Doctor AI system, which is a RNN-based model that can learn efficient patient representation from a large amount of longitidinal patient records and predict future events of patients. We tested Doctor AI on a large real-world EHR datasets, which achieved 79.58\% recall@30 and significantly outperformed many baselines. 
We have also shown that the patient's visit count and the rarity of medical codes highly influence the performance.
%longer patient records improve the prediction accuracy, and that our model has a hard time learning from rarely occurring codes such as Kleinfelter's syndrome. 
We have also demonstrated that knowledge learned from one hospital could be adapted to another hospital.
The empirical analysis by a medical expert confirmed that Doctor AI not only mimics the predictive power of human doctors, but also provides diagnostic results that are clinically meaningful.% in that all the predicted codes were within the boundary of medical possibility, indicating its potential to be used as a differential diagnosis machine.

One limitation of Doctor AI is that, in medical practice, incorrect predictions can sometimes be more important than correct predictions as they can degrade patient health. Also, although Doctor AI has shown that it can mimic physicians' average behavior, it would be more useful to learn to perform better than average. We set as our future work to address these issues so that Doctor AI can provide practical help to physicians in the future.
%Our proposed Doctor AI can potentially open up new avenues for further improvement of computational health problems. Applications and extensions of Doctor AI can be studied in diverse clinical settings including expanding toward rich and unstructured data sources such as medical images and clinical notes.

% ACKNOWLEDGEMENTS ONLY GO IN THE CAMERA-READY, NOT THE SUBMISSION
% \acks{Many thanks to all collaborators and funders!}

\bibliography{references}

\section*{Appendices}
\appendix
\section{Description of Gated Recurrent Units}
\label{sec:gru}
\begin{figure}[!h]
\centering
\includegraphics[scale=0.3]{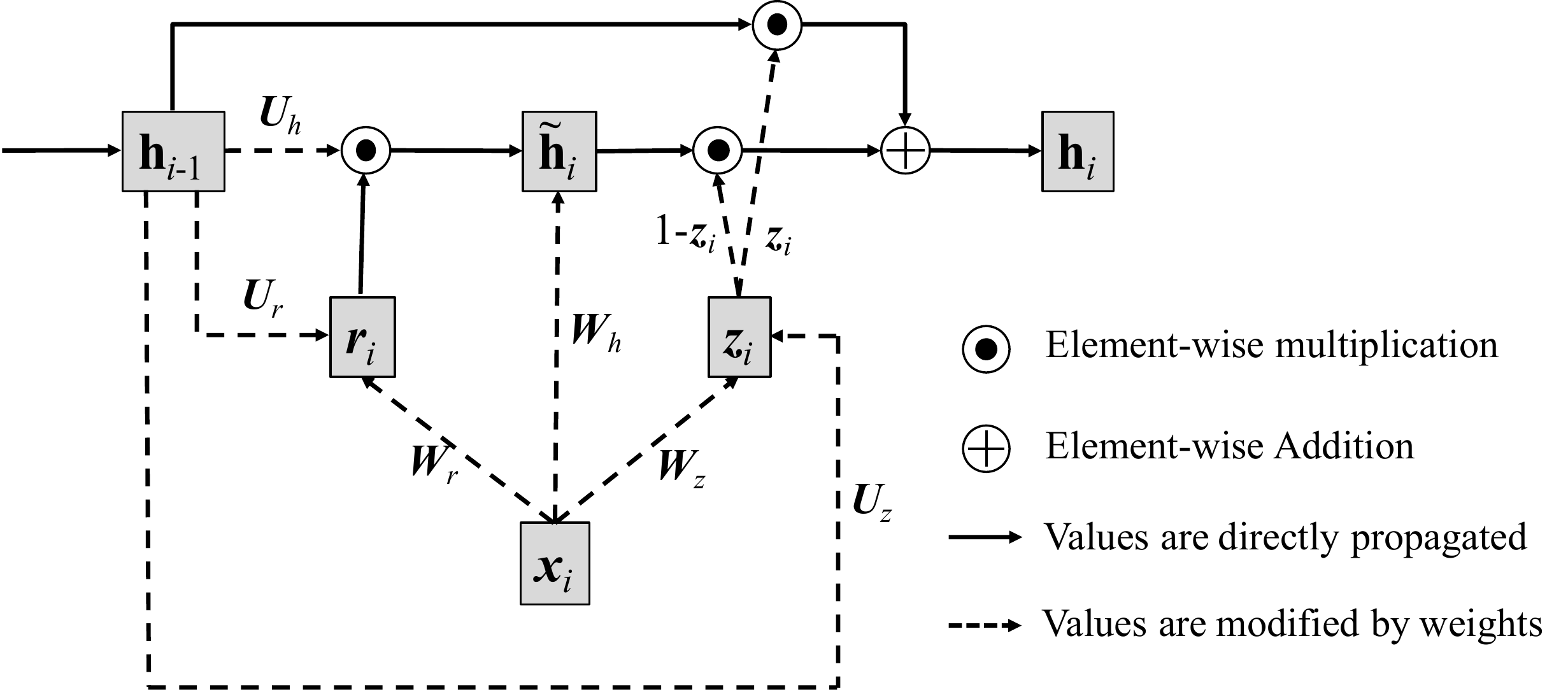}
\caption{Architecture of GRU}
\label{fig:gru}
\end{figure}

\noindent We first reiterate the mathematical formulation of GRU so that the reader can see Figure \ref{fig:gru} and the formulations together.
\begin{align*}
\zb_i & = \sigma(\Wb_z \xb_i + \Ub_z \hb_{i-1} + \bb_z) \\
\rb_i & = \sigma(\Wb_r \xb_i + \Ub_r \hb_{i-1} + \bb_r) \\
\tilde{\hb}_i & = \tanh(\Wb_h \xb_i + \rb_i \circ \Ub_h \hb_{i-1} + \bb_h) \\
\hb_i & = \zb_i \circ \hb_{i-1} + (1 - \zb_i) \circ \tilde{\hb}_i
\end{align*}
Figure \ref{fig:gru} depicts the architecture of the GRU, where $\xb_i$, $\zb_i$ and $\rb_i$ respectively represent the input, update gate and the reset gate, $\tilde{\hb}_i$ the intermediate memory unit, $\hb_i$ the hidden layer, all at timestep $t_i$. $\Wb_h, \Wb_z, \Wb_r, \Ub_h, \Ub_z, \Ub_r$ are the weight matrices to be learned. Note that the bias vectors $\bb_h, \bb_z, \bb_r$ are omitted in Figure \ref{fig:gru}.

The outstanding difference between the classical RNN (Elman Network) and GRU is that the previous hidden layer $\hb_{i-1}$ and the current input $\xb_i$ do not directly change the value of the current hidden layer $\hb_i$. Instead, they change the values of both gates $\zb_i$, $\rb_i$ and the intermediate memory unit $\tilde{\hb}_i$. Then the current hidden layer $\hb_i$ is updated by $\tilde{\hb}_i$ and $\zb_i$. Due to the $\sigma$ function, both gates $\zb_i$ and $\rb_i$ have values between 0 and 1. Therefore if the reset gate $\rb_i$ is close to zero, the intermediate memory unit $\tilde{\hb}_i$ will disregard the past values of the hidden layer $\hb_{i-1}$. If the update gate $\zb_i$ is close to one, the current hidden layer $\hb_i$ will disregard the current input $\xb_i$, and retain the value from the previous timestep $\hb_{i-1}$.

Simply put, the reset gate allows the hidden layer to drop any information that is not useful in making a prediction, and the updated gate controls how much information from the previous hidden layer should be propagated to the current hidden layer. This characteristic of GRU is especially useful as it is not easy to identify information essential to predicting the future diagnosis, medication or the time duration until the next visit.

\section{Learning the Skip-gram vectors from the EHR}
\label{sec:skipgram}
Learning efficient representations of medical codes (\textit{e.g.} diagnosis codes, medication codes, and procedure codes) may lead to improved performance of many clinical applications. We specifically used Skip-gram \cite{mikolov2013distributed} to learn real-valued multidimensional vectors to capture the latent representation of medical codes from the EHR. 

We processed the private dataset so that diagnosis codes, medication codes, procedure codes are laid out in a temporal order. If there are multiple codes at a single visit, they were laid out in a random order. Then using the context window size of 5 to the left and 5 to the right, and applying Skip-gram, we were able to project diagnosis codes, medication codes and procedure codes into the same lower dimensional space, where similar or related codes are embedded close to one another. For example, hypertension, obesity, hyperlipidemia all share similar values compared to pneumonia or bronchitis. The trained Skip-gram vectors are then plugged into RNN so that a multi-hot vector can be converted to vector representations of medical codes. 

\section{Details of the training procedure of multilayer perceptron}
\label{sec:base}
%\ecedit{As mentioned in \ref{sec:data}, we use HPOlib, a bayesian optimization tool to tune the hyperparameters for the baseline models, specifically logistic regression and multilayer perceptron (MLP). $75\%$ of the patients were used as the training set, and $10\%$ of the patients were used as the validation set. For logistic regression, we tune the $L_2$ regularization coefficients (0.0001$\sim$10000.0) for all weight matrices. For multilayer perceptron, we tune the number of hidden layers ($1,2,3$), the number of neurons in the hidden layers (100$\sim$2000), the dropout rate ($p=$ 0.0$\sim$0.9), and the $L_2$ regularization coefficients (0.0001$\sim$10000.0) for all weight matrices. For logistic regressionThe activation functions in the hidden layers and the output layer are set to tanh and softmax functions respectively. For prediction of time intervals, we use rectified linear units for both logistic regression and MLP. Once the optimal hyperparameters are chosen, we train the optimal model using both the training set and the validation set. Then we evaluate the performance against the remaining $15\%$ of the patients. The results are recorded in Table \ref{tab:mainResults}

We use a multilayer perceptron with a hidden layer of width 2,000. We apply $L_2$ regularization to all of the weight matrices. The activation functions in the first and output layers are selected to be tanh and softmax functions respectively. For prediction of time intervals, we used rectified linear units.

%\subsection{Manual feature extraction details}
%\label{sec:manFeature}
%The process of extracting \textit{tf-idf} like features for each visit is as follows:  First, we compute the $p$-dimensional count vector  by counting the number of visits in which each code has been observed. The \textit{idf} vector is computing by taking the logarithm of total number of visits divided by the count vector.
%
%Then, the \textit{tf} vector is computed for each visit by counting the number of codes observed in \textit{all visits} until the current visit.  This process is used to ensure that the features include information from past visits.  Finally, we use the element-wise product of the \textit{tf} and \textit{idf} vectors as the \textit{tf-idf} vector.

\section{Case study}
\label{sec:qual}
The detailed results are shown in Table \ref{tab:qual}.
To take a closer look at the performance of Doctor AI, in Table \ref{tab:qual} (in Appendix \ref{sec:qual}) we list the predicted, true, and historical diagnosis codes for five visits of different patients. The blue items represent the correct predictions. The results are promising and show that, given the history of the patient, the Doctor AI can predict the true diagnostic codes. The results highly mimic the way a human doctor will interpret the disease predictions from the history. For all five of the cases shown in Table \ref{tab:qual}, the set of predicted diseases contain most, if not all of the true diseases. For example, in the first case, the top 3 predicted diseases match the true diseases. A human doctor would likely predict similar diseases to the ones predicted with Doctor AI, since old myocardial infarction and chronic ischemic heart disease can be associated with infections and diabetes \citep{stevens1978diabetic}.

In the fourth case, visual disturbances can be associated with migraines and essential hypertension \citep{keith1939some}. Further, essential hypertension may be linked to cognitive function \citep{kuusisto1993essential}, which plays a role in anxiety disorders and dissociative and somatoform disorders. Regarding codes that are guessed incorrectly with the fourth case, they can still be plausible given the history. For example, cataracts, and disorders of refraction and accommodation could have been guessed based on a history of visual disturbances, as well as strabismus and disorders of binocular eye movements. Allergic rhinitis could have been guessed, because there was a history of allergic rhinitis. In summary, Doctor AI is able to very accurately predict the true diagnoses in the sample patients. The results are promising and should motivate future studies involving the application of Doctor AI on different datasets exhibiting other populations of patients.

%\subsection{Supplementary clinical explanation} \label{sec:robert}

\newgeometry{left=1.5cm,bottom=0.5cm} 
\thispagestyle{empty}
\begin{landscape}
\setlength\LTcapwidth{\textwidth} % default: 4in (rather less than \textwidth...)
\setlength\LTleft{0pt}            % default: \parindent
\setlength\LTright{0pt}           % default: \fill

\begin{tiny}
\begin{table}[t]
\scriptsize
\centering
\caption{Comparison of the diagnoses by Doctor AI and the true future diagnoses.}
\label{tab:qual}
\begin{tabular}{|@{}c@{\;}|@{\;}l@{}|@{}c@{\;}|@{\;}l@{}|@{}c@{\;}|@{\;}l@{\;}|}
\hline
\multicolumn{2}{|c|}{\textbf{Predicted}}
& \multicolumn{2}{c|}{\textbf{True}}
& \multicolumn{2}{c|}{\textbf{History}}
\\ \hline
ICD9   & Description & ICD9   & Description & ICD9  & Description \\ 
\hline\hline
\begin{tabular}[c]{@{}c@{}}\blue{412}\\ \blue{V58}\\ \blue{414}\\ 272\\ 250\\ 585\\ 428\\ 285\\ V04\\ V76\\ \end{tabular} &
 \begin{tabular}[c]{@{}l@{}}\blue{Old myocardial infarction}\\ \blue{Encounter for other and unspecified procedures}\\ \blue{Other forms of chronic ischemic heart disease}\\ Disorders of lipoid metabolism\\ Diabetes mellitus\\ Chronic kidney disease (CKD)\\ Heart failure\\ Other and unspecified anemias\\ \scriptsize{Need for prophylactic vaccin. and inocul. against certain diseases}\\ Special screening for malignant neoplasms\\ \end{tabular} 
&\begin{tabular}[c]{@{}c@{}}\blue{414}\\ \blue{412}\\ \blue{V58}\\ \end{tabular} &
 \begin{tabular}[c]{@{}l@{}}\blue{Other forms of chronic ischemic heart disease}\\ \blue{Old myocardial infarction}\\ \blue{Encounter for other and unspecified procedures}\\ \end{tabular} 
&\begin{tabular}[c]{@{}c@{}}465\\ 250\\ 366\\ V58\\ 362\\ \end{tabular} &
 \begin{tabular}[c]{@{}l@{}}\scriptsize{Acute upper respiratory infec. of multiple or unspec. sites}\\ Diabetes mellitus\\ Cataract\\ Encounter for other and unspecified procedures\\ Other retinal disorders\\ \end{tabular} 
\\ 
 \hline 
\begin{tabular}[c]{@{}c@{}}\blue{V07}\\ 477\\ 780\\ \blue{401}\\ \blue{786}\\ 493\\ 300\\ 461\\ 530\\ 719\\ \end{tabular} &
 \begin{tabular}[c]{@{}l@{}}\blue{Need for isolation and other prophylactic measures}\\ Allergic rhinitis\\ General symptoms\\ \blue{Essential hypertension}\\ \blue{\scriptsize{Symptoms involving respiratory system}}\\ Asthma\\ Anxiety, dissociative and somatoform disorders\\ Acute sinusitis\\ Diseases of esophagus\\ Other and unspecified disorders of joint\\ \end{tabular} 
&\begin{tabular}[c]{@{}c@{}}\blue{V07}\\ \blue{401}\\ \blue{786}\\ 782\\ \end{tabular} &
 \begin{tabular}[c]{@{}l@{}}\blue{Need for isolation and other prophylactic measures}\\ \blue{Essential hypertension}\\ \blue{\scriptsize{Symptoms involving respiratory system}}\\ Symptoms involving skin and other integumentary tissue\\ \end{tabular} 
&\begin{tabular}[c]{@{}c@{}}782\\ 477\\ V07\\ 564\\ 401\\ \end{tabular} &
 \begin{tabular}[c]{@{}l@{}}Symptoms involving skin and other integumentary tissue\\ Allergic rhinitis\\ Need for isolation and other prophylactic measures\\ Functional digestive disorders, not elsewhere classified\\ Essential hypertension\\ \end{tabular} 
\\ 
 \hline 
\begin{tabular}[c]{@{}c@{}}453\\ \blue{V58}\\ \blue{719}\\ \blue{V12}\\ V43\\ 729\\ \blue{715}\\ 733\\ 726\\ 451\\ \end{tabular} &
 \begin{tabular}[c]{@{}l@{}}Other venous embolism and thrombosis\\ \blue{Encounter for other and unspecified procedures}\\ \blue{Other and unspecified disorders of joint}\\ \blue{Personal history of certain other diseases}\\ Organ or tissue replaced by other means\\ Other disorders of soft tissues\\ \blue{Osteoarthrosis and allied disorders}\\ Other disorders of bone and cartilage\\ Peripheral enthesopathies and allied syndromes\\ Phlebitis and thrombophlebitis\\ \end{tabular} 
&\begin{tabular}[c]{@{}c@{}}\blue{715}\\ \blue{V12}\\ \blue{719}\\ \blue{V58}\\ \end{tabular} &
 \begin{tabular}[c]{@{}l@{}}\blue{Osteoarthrosis and allied disorders}\\ \blue{Personal history of certain other diseases}\\ \blue{Other and unspecified disorders of joint}\\ \blue{Encounter for other and unspecified procedures}\\ \end{tabular} 
&\begin{tabular}[c]{@{}c@{}}453\\ 956\\ V43\\ \end{tabular} &
 \begin{tabular}[c]{@{}l@{}}Other venous embolism and thrombosis\\ Injury to peripheral nerve(s) of pelvic girdle and lower limb\\ Organ or tissue replaced by other means\\ \end{tabular} 
\\ 
 \hline 
\begin{tabular}[c]{@{}c@{}}477\\ \blue{780}\\ \blue{300}\\ \blue{401}\\ \blue{346}\\ 366\\ V43\\ 367\\ 368\\ 272\\ \end{tabular} &
 \begin{tabular}[c]{@{}l@{}}Allergic rhinitis\\ \blue{General symptoms}\\ \blue{Anxiety, dissociative and somatoform disorders}\\ \blue{Essential hypertension}\\ \blue{Migraine}\\ Cataract\\ Organ or tissue replaced by other means\\ Disorders of refraction and accommodation\\ Visual disturbances\\ Disorders of lipoid metabolism\\ \end{tabular} 
&\begin{tabular}[c]{@{}c@{}}\blue{401}\\ \blue{780}\\ \blue{346}\\ \blue{300}\\ \end{tabular} &
 \begin{tabular}[c]{@{}l@{}}\blue{Essential hypertension}\\ \blue{General symptoms}\\ \blue{Migraine}\\ \blue{Anxiety, dissociative and somatoform disorders}\\ \end{tabular} 
&\begin{tabular}[c]{@{}c@{}}782\\ 477\\ 692\\ 368\\ 378\\ \end{tabular} &
 \begin{tabular}[c]{@{}l@{}}Symptoms involving skin and other integumentary tissue\\ Allergic rhinitis\\ Contact dermatitis and other eczema\\ Visual disturbances\\ Strabismus and other disorders of binocular eye movements\\ \end{tabular} 
\\ 
 \hline 
\begin{tabular}[c]{@{}c@{}}\blue{428}\\ \blue{427}\\ \blue{272}\\ 401\\ 786\\ 185\\ \blue{250}\\ 414\\ 788\\ 424\\ \end{tabular} &
 \begin{tabular}[c]{@{}l@{}}\blue{Heart failure}\\ \blue{Cardiac dysrhythmias}\\ \blue{Disorders of lipoid metabolism}\\ Essential hypertension\\ \scriptsize{Symptoms involving respiratory system}\\ Malignant neoplasm of prostate\\ \blue{Diabetes mellitus}\\ Other forms of chronic ischemic heart disease\\ Symptoms involving urinary system\\ Other diseases of endocardium\\ \end{tabular} 
&\begin{tabular}[c]{@{}c@{}}\blue{250}\\ 402\\ \blue{428}\\ \blue{272}\\ \blue{427}\\ \end{tabular} &
 \begin{tabular}[c]{@{}l@{}}\blue{Diabetes mellitus}\\ Hypertensive heart disease\\ \blue{Heart failure}\\ \blue{Disorders of lipoid metabolism}\\ \blue{Cardiac dysrhythmias}\\ \end{tabular} 
&\begin{tabular}[c]{@{}c@{}}466\\ 428\\ 786\\ 785\\ 250\\ \end{tabular} &
 \begin{tabular}[c]{@{}l@{}}Acute bronchitis and bronchiolitis\\ Heart failure\\ \scriptsize{Symptoms involving respiratory system}\\ Symptoms involving cardiovascular system\\ Diabetes mellitus\\ \end{tabular} 
\\ 
 \hline 
\end{tabular}
\end{table}
\end{tiny}

\end{landscape}

%%  Second page
\thispagestyle{empty}
\begin{landscape}
\setlength\LTcapwidth{\textwidth} % default: 4in (rather less than \textwidth...)
\setlength\LTleft{0pt}            % default: \parindent
\setlength\LTright{0pt}           % default: \fill

\begin{tiny}
\begin{table}[t]
\scriptsize
\centering
\caption{Comparison of the diagnoses by Doctor AI for a frequent and an infrequent disease code after 200 time step.}
\label{tab:visu}
\begin{tabular}{|c|l||c|l|}
\hline
\multicolumn{2}{|c||}{\textbf{Hypertension}}
& \multicolumn{2}{c|}{\textbf{Klinefelter's syndrome}}
\\ \hline
ICD9   & Description & ICD9   & Description  \\ 
\hline\hline
401 &	Essential hypertension &  272 &	Disorders of lipoid metabolism \\
272 &	Disorders of lipoid metabolism &  V70 &	General medical examination \\
786 &	Symptoms involving respiratory system and other chest symptoms &  V04 &	Need for prophylactic vaccination and inoculation against certain diseases \\
V06 &	Need for prophylactic vaccination and inoculation against combinations of diseases &  730 &	Osteomyelitis, periostitis, and other infections involving bone \\
790 &	Nonspecific findings on examination of blood &  780 &	General symptoms \\
V76 &	Special screening for malignant neoplasms &  783 &	Symptoms concerning nutrition, metabolism, and development \\
V04 &	Need for prophylactic vaccination and inoculation against certain diseases &  295 &	Schizophrenic disorders \\
V70 &	General medical examination &  V76 &	Special screening for malignant neoplasms \\
780 &	General symptoms &  141 &	Malignant neoplasm of tongue \\
276 &	Disorders of fluid, electrolyte, and acid-base balance &  V06 &	Need for prophylactic vaccination and inoculation against combinations of diseases \\
782 &	Symptoms involving skin and other integumentary tissue &  250 &	Diabetes mellitus \\
268 &	Vitamin D deficiency &  782 &	Symptoms involving skin and other integumentary tissue \\
719 &	Other and unspecified disorders of joint &  786 &	Symptoms involving respiratory system and other chest symptoms \\
427 &	Cardiac dysrhythmias &  208 &	Leukemia of unspecified cell type \\
380 &	Disorders of external ear &  401 &	Essential hypertension \\
250 &	Diabetes mellitus &  790 &	Nonspecific findings on examination of blood \\
599 &	Other disorders of urethra and urinary tract &  280 &	Iron deficiency anemias \\
V72 &	Special investigations and examinations &  607 &	Disorders of penis \\
789 &	Other symptoms involving abdomen and pelvis &  281 &	Other deficiency anemias \\
729 &	Other disorders of soft tissues &  V03 &	Need for prophylactic vaccination and inoculation against bacterial diseases \\
682 &	Other cellulitis and abscess &  332 &	Parkinson's disease \\
V03 &	Need for prophylactic vaccination and inoculation against bacterial diseases &  255 &	Disorders of adrenal glands \\
724 &	Other and unspecified disorders of back &  799 &	Other ill-defined and unknown causes of morbidity and mortality \\
V58 &	Encounter for other and unspecified procedures and aftercare &  244 &	Acquired hypothyroidism \\
278 &	Overweight, obesity and other hyperalimentation &  V58 &	Encounter for other and unspecified procedures and aftercare \\
V82 &	Special screening for other conditions &  151 &	Malignant neoplasm of stomach \\
V65 &	Other persons seeking consultation &  294 &	Persistent mental disorders due to conditions classified elsewhere \\
585 &	Chronic kidney disease (CKD) &  V72 &	Special investigations and examinations \\
274 &	Gout &  344 &	Other paralytic syndromes \\
V49 &	Other conditions influencing health status &  146 &	Malignant neoplasm of oropharynx \\
\hline
\end{tabular}
\end{table}
\end{tiny}

\end{landscape}
\restoregeometry 

\end{document}